\documentclass[runningheads]{llncs}
\usepackage{graphicx}
\usepackage{hyperref}
\usepackage{url}
\usepackage{float}
\usepackage{caption}
\usepackage{subcaption}
\usepackage{algorithm}
\usepackage[noend]{algpseudocode}
\usepackage[bottom]{footmisc}
\usepackage[utf8]{inputenc} 
\usepackage[T1]{fontenc}    
\usepackage{booktabs}       
\usepackage{amsfonts}       
\usepackage{nicefrac}       
\usepackage{microtype}      

\usepackage{amsmath}
\usepackage{xcolor }
\usepackage{amssymb}

\begin{document}
\title{TextKD-GAN: Text Generation using Knowledge Distillation and Generative Adversarial Networks}

\author{
  Md. Akmal Haidar \and Mehdi Rezagholizadeh \\
 \email{\{md.akmal.haidar, mehdi.rezagholizadeh\}@huawei.com}
  }

\institute{Huawei Noah's Ark Lab, Montreal Research Center, Montreal, Canada}

%
\maketitle              
\begin{abstract}
Text generation is of particular interest in many NLP applications such as  machine translation, language modeling, and text summarization. Generative adversarial networks (GANs) achieved a remarkable success in high quality image generation in computer vision, and recently, GANs have gained lots of interest from the NLP community as well. However, achieving similar success in NLP would be more challenging due to the discrete nature of text. In this work, we introduce a method using knowledge distillation to effectively exploit GAN setup for text generation. We demonstrate how autoencoders (AEs) can be used for providing a continuous representation of sentences, which is a smooth representation that assign non-zero probabilities to more than one word. We distill this representation to train the generator to synthesize similar smooth representations. We perform a number of experiments to validate our idea using different datasets and show that our proposed approach yields better performance in terms of the BLEU score and Jensen-Shannon distance (JSD) measure compared to traditional GAN-based text generation approaches without pre-training.

\keywords{Text generation  \and Generative adversarial networks \and Knowledge distillation.}
\end{abstract}

\section{Introduction}

Recurrent neural network (RNN) based techniques such as language models are the most popular approaches for text generation. These RNN-based text generators rely on maximum likelihood estimation (MLE) solutions such as teacher forcing~\cite{williams1989} (i.e. the model is trained to predict the next item given all previous observations); however, it is  well-known in the literature that MLE is a simplistic objective for this complex NLP task~\cite{li2017adversarial}. MLE-based methods suffer from exposure bias~\cite{rajeswar2017adversarial}, which means that at training time the model is exposed to gold data only, but at test time it observes its own predictions.

However, GANs which are based on the adversarial loss function and have the generator and the discriminator networks suffers less from the mentioned problems. GANs could provide a better image generation framework comparing to the traditional MLE-based methods and achieved substantial success in the field of computer vision for generating realistic and sharp images. This great success motivated researchers to apply its framework to NLP applications as well. 

GANs have been exploited recently in various NLP applications such as machine translation \cite{wu2017adversarial,yang2017improving}, dialogue models~\cite{li2017adversarial}, question answering~\cite{yang2017semi}, and natural language generation~\cite{gulrajani2017improved,rajeswar2017adversarial,press2017language,kim2017adversarially,zhang2017adversarial,Zhu2018}. 
However, applying GAN in NLP is challenging due to the discrete nature of the text. Consequently, back-propagation would not be feasible for discrete outputs and it is not straightforward to pass the gradients through the discrete output words of the generator. The existing GAN-based solutions can be categorized according to the technique that they leveraged for handling the problem of the discrete nature of text: Reinforcement learning (RL) based methods, latent space based solutions, and approaches based on continuous approximation of discrete sampling.  
Several versions of the RL-based techniques have been introduced in the literature including Seq-GAN~\cite{yu2017seqgan}, MaskGAN~\cite{fedus2018maskgan}, and LeakGAN~\cite{guo2017long}.
However, they often need pre-training and are computationally more expensive compared to the methods of the other two categories.
Latent space-based solutions derive a latent space representation of the text using an AE and attempt to learn data manifold of that space \cite{kim2017adversarially}.  
Another approach for generating text with GANs is to find a continuous approximation of the discrete sampling by using the Gumbel Softmax technique~\cite{kusner2016gans} or approximating the non-differentiable argmax operator~\cite{zhang2017adversarial} with a continuous function. 

In this work, we introduce TextKD-GAN as a new solution for the main bottleneck of using GAN for text generation with knowledge distillation: a technique that transfer the knowledge of softened output of a teacher model to a student model~\cite{Hinton2015}. Our solution is based on an AE (Teacher) to derive a smooth representation of the real text.  This smooth representation is fed to the TextKD-GAN discriminator instead of the conventional one-hot representation. The generator (Student) tries to learn the manifold of the softened smooth representation of the AE. We show that TextKD-GAN outperforms the conventional GAN-based text generators that do not need pre-training. 
The remainder of the paper is organized as follows.  In the next two sections, some preliminary background on generative adversarial networks and related work in the literature will be reviewed. The proposed method will be presented in section~\ref{methodology}. In section~\ref{results}, the experimental details will be discussed. Finally, section~\ref{conclusion} will conclude the paper.

\section{Background}

Generative adversarial networks include two separate deep networks: a generator and a discriminator. The generator takes in a random variable, $z$ following a distribution $P_z (z)$ and attempt to map it to the data distribution $P_x(x)$. The output distribution of the generator is expected to converge to the data distribution during the training. On the other hand, the discriminator is expected to discern real samples from generated ones by outputting zeros and ones, respectively.
During training, the generator and discriminator generate samples and classify them, respectively by adversarially affecting the performance of each other. In this regard, an adversarial loss function is employed for training~\cite{goodfellow2014generative}: 
\begin{equation}
\label{eq1}
\begin{split}
\min_G \max_D V(D, G) = E_{x \sim P_x(x)} [\text{log} D(x)]+E_{z\sim P_z(z)}[\text{log}(1-D(G(z)))]  
\end{split}
\end{equation} 
This is a two-player minimax game for which a Nash-equilibrium point should be derived. Finding the solution of this game is non-trivial and there has been a great extent of literature dedicated in this regard~\cite{salimans2016improved}. 
\begin{figure}[H]
\centering
\includegraphics[scale=0.40]{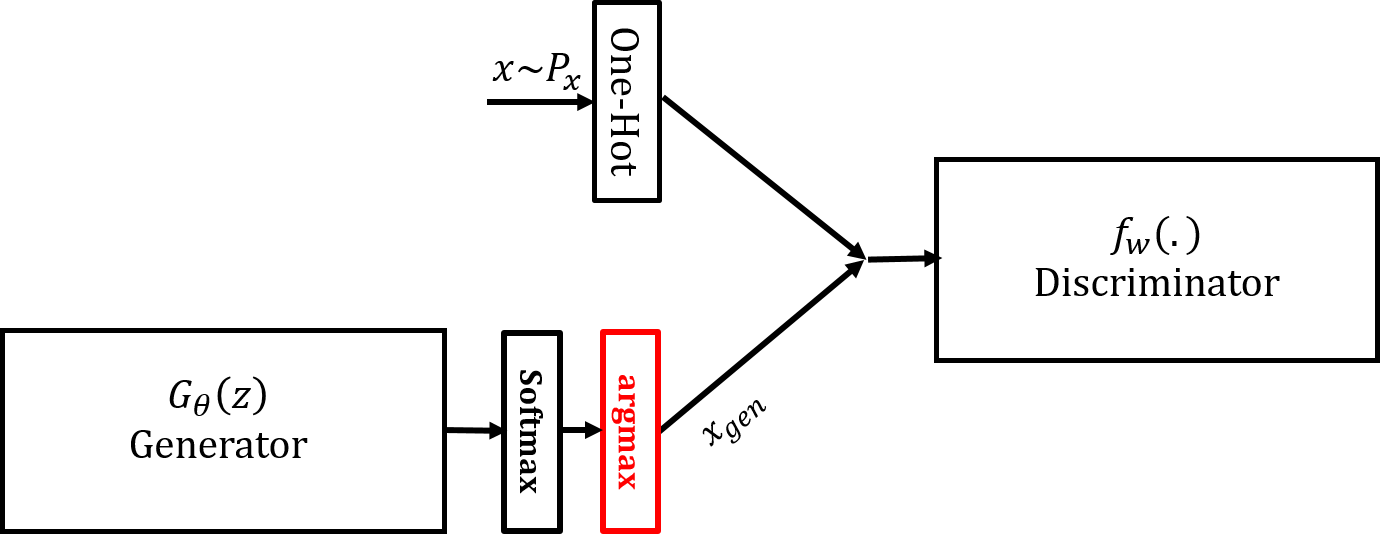}
\caption{Simplistic text generator with GAN}
\label{simple_GAN}
\end{figure}

As stated, using GANs for text generation is challenging because of the discrete nature of text. To clarify the issue, Figure~\ref{simple_GAN} depicts a simplistic architecture for GAN-based text generation. The main bottleneck of the design is the \textit{argmax} operator which is not differentiable and blocks the gradient flow from the discriminator to the generator.

\begin{equation}
\begin{split}
& \min_G  E_{z\sim P_z(z)}[\text{log}(1-D(\text{argmax}(\text{softmax}(G(z)))))]  
\end{split}
\end{equation}

\subsection{Knowledge Distillation}
Knowledge distillation has been studied in model compression where knowledge of a large cumbersome model is transferred to a small model for easy deployment. Several studies have been studied on the knowledge transfer technique~\cite{Hinton2015,Romero2015}. It starts by training a big teacher model (or ensemble model) and then train a small student model which tries to mimic the characteristics of the teacher model, such as hidden representations~\cite{Romero2015}, it's output probabilities~\cite{Hinton2015}, or directly on the generated sentences by the teacher model in neural machine translation~\cite{Kim2016}. The first teacher-student framework for knowledge distillation was proposed in~\cite{Hinton2015} by introducing the softened teacher's output. In this paper, we propose a GAN framework for text generation where the generator (Student) tries to mimic the reconstructed output representation of an auto-encoder (Teacher) instead of mapping to a conventional one-hot representations.  

\subsection{Improved WGAN}
\begin{figure}[!htb]
\centering
\includegraphics[scale=0.40]{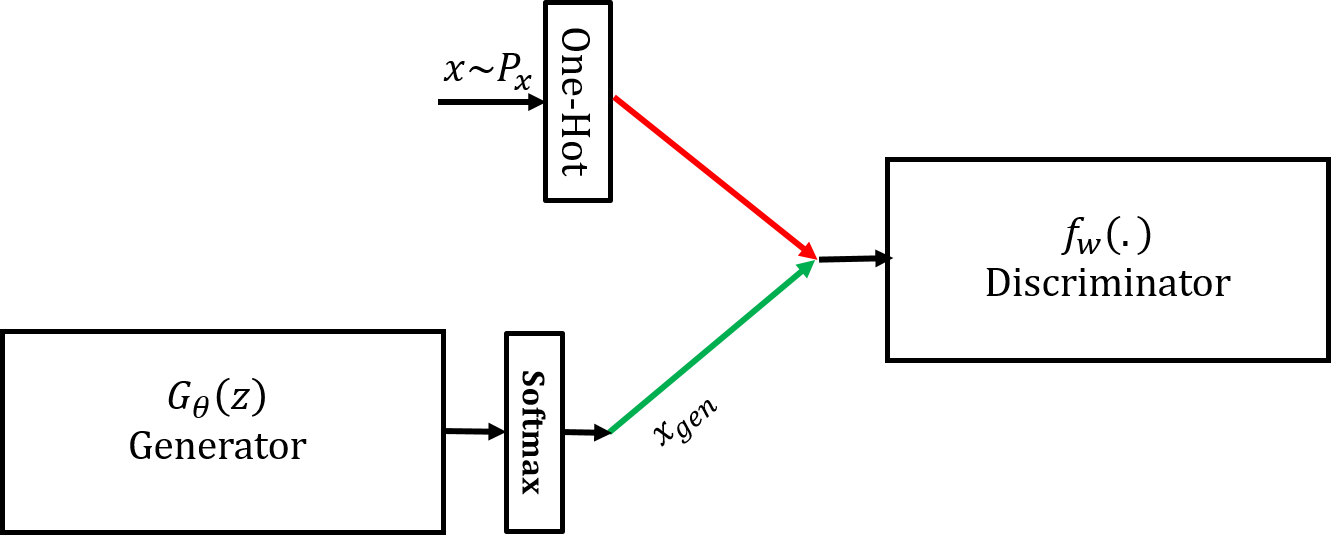}
\caption{Improved WGAN for text generation}
\label{Fig_IWGAN}
\end{figure}

Generating text with pure GANs is inspired by improved Wasserstein GAN (IWGAN) work~\cite{gulrajani2017improved}. In IWGAN, a character level language model is developed based on adversarial training of a generator and a discriminator without using any extra element such as  policy gradient reinforcement learning~\cite{Sutton1999}. The generator produces a softmax vector over the entire vocabulary. The discriminator is responsible for distinguishing between the one-hot representations of the real text and the softmax vector of the generated text. The IWGAN method is described in Figure~\ref{Fig_IWGAN}. A disadvantage of this technique is that the discriminator is able to tell apart the one-hot input from the softmax input very easily. Hence, the generator will have a hard time fooling the discriminator and vanishing gradient problem is highly probable. 
   
\section{Related Work}
A new version of Wasserstein GAN for text generation using gradient penalty for discriminator was proposed in~\cite{gulrajani2017improved}. Their generator is a CNN network generating fixed-length texts. The discriminator is another CNN receiving 3D tensors as input sentences. It determines whether the tensor is coming from the generator or sampled from the real data. The real sentences and the generated ones are represented using one-hot and softmax representations, respectively.

A similar approach was proposed in~\cite{rajeswar2017adversarial} with an RNN-based generator. They used a curriculum learning strategy~\cite{bengio2009curriculum} to produce sequences of gradually increasing lengths as training progresses. 
In~\cite{press2017language}, RNN is trained to generate text with GAN using curriculum learning. The authors proposed a procedure called teacher helping, which helps the generator to produce long sequences by conditioning on shorter ground-truth sequences. 

All these approaches use a discriminator to discriminate the generated softmax output from one-hot real data as in Figure~\ref{Fig_IWGAN}, which is a clear downside for them. The reason is the discriminator receives inputs of different representations: a one-hot vector for real data and a probabilistic vector output from the generator. It makes the discrimination rather trivial.  

AEs have been exploited along with GANs in different architectures for computer vision application such as AAE~\cite{makhzani2015adversarial}, ALI~\cite{dumoulin2016adversarially}, and HALI~\cite{belghazi2018hierarchical}. 
Similarly, AEs can be used with GANs for generating text. For instance, an adversarially regularized AE (ARAE) was proposed  in~\cite{kim2017adversarially}. The generator is trained in parallel to an AE to learn a continuous version of the code space produced by AE encoder. Then, a discriminator will be responsible for distinguishing between the encoded hidden 
code and the continuous code of the generator. Basically, in this approach, a continuous distribution is generated corresponding to an encoded code of text.

\section{Methodology}
\label{methodology}
AEs can be useful in denoising text and transferring it to a code space (encoding) and then reconstructing back to the original text from the code. AEs can be combined with GANs in order to improve the generated text. In this section, we introduce a technique using AEs to replace the conventional one-hot representation ~\cite{gulrajani2017improved} with a continuous softmax representation of real data for discrimination. 

\subsection{Distilling output probabilities of AE to TextKD-GAN generator}

\begin{figure}[H]
\centering
\includegraphics[scale=0.4]{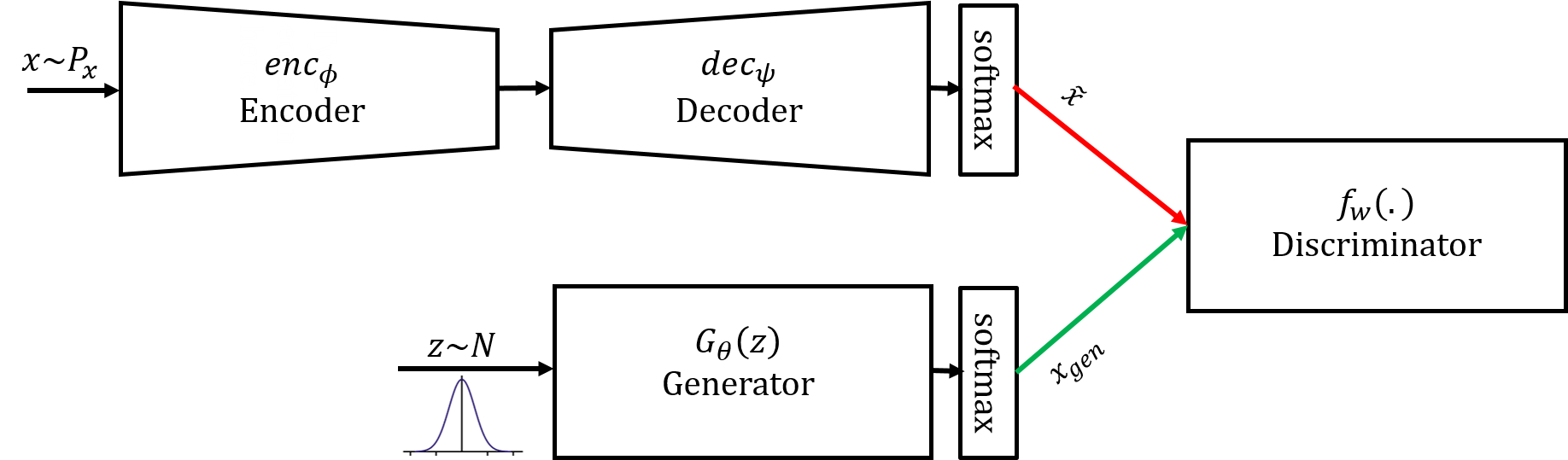}
\caption{TextKD-GAN model for text generation}
\label{Fig_AE_GAN}
\end{figure}

As stated, in conventional text-based discrimination approach~\cite{gulrajani2017improved}, the real and generated input of the discriminator will have different types (one-hot and softmax) and it can simply tell them apart. One way to avoid this issue is to derive a continuous smooth  representation of words rather than their one-hot and train the discriminator to differentiate between the continuous representations.
In this work, we use a conventional AE (Teacher) to replace the one-hot representation with softmax reconstructed output, which is a smooth representation that yields smaller variance in gradients~\cite{Hinton2015}. The proposed model is depicted in Figure~\ref{Fig_AE_GAN}. As seen, instead of the one-hot representation of the real words, we feed the softened  reconstructed output of the AE to the discriminator.
This technique would makes the discrimination much harder for the discriminator. The GAN generator (Student) with softmax output tries to mimic the AE output distribution instead of conventional one-hot representations used in the literature.   

\subsection{Why TextKD-GAN should Work Better than IWGAN}
\begin{figure}[!htb]
\begin{center}
\fbox{
\begin{minipage}[t]{.4\linewidth}
\centering
\includegraphics[scale=0.7]{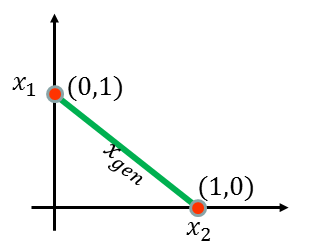}
\end{minipage}}
\fbox{
\begin{minipage}[t]{.4\linewidth}
\centering
\includegraphics[scale=0.672]{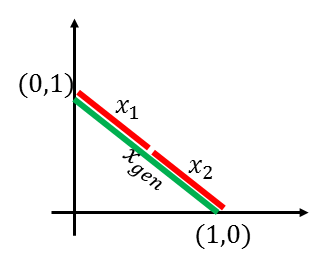}
\end{minipage}}
\end{center}
\caption{Locus of  the input vectors to the discriminator for a two-word language model; Left panel: IWGAN, Right panel: TextKD-GAN.}
\label{Fig_ex}
\end{figure}

Suppose we apply IWGAN to a language vocabulary of size two: words $x_1$ and $x_2$. The one-hot representation of these two words (as two points in the Cartesian coordinates) and the span of the generated softmax outputs (as a line segment connecting them) is depicted in the left panel of Figure~\ref{Fig_ex}.
As evident graphically, the task of the discriminator is to discriminate the points from the line connecting them, which is a rather simple very easy task. 

Now, let's consider the TextKD-GAN idea using the two-word language example. As depicted in Figure ~\ref{Fig_ex} (Right panel), the output locus of the TextKD-GAN decoder would be two red line segments instead of two points (in the one-hot case). The two line segments lie on the output locus of the generator, which will make the generator more successful in fooling the discriminator.

\subsection{Model Training}
We train the AE and TextKD-GAN simultaneously. In order to do so, we break down the objective function into three terms: 
(1) a reconstruction term for the AE, (2) a discriminator loss function with gradient penalty, (3) an adversarial cost for the generator. Mathematically,
\begin{equation}
\begin{split}
& 1) \min_{(\phi,\psi)} L_{AE} (\phi,\psi) = \min_{(\phi,\psi)} ||x-\text{softmax}(\text{dec}_\psi(\text{enc}_\phi(x)))||^2 \\
& 2) \min_{w \in W}  L_{\text{discriminator}} (w) =\\
& \min_{w \in W} - E_{x \sim P_x} [f_w (\text{dec}_{\psi} (\text{enc}_{\phi} (X)))]+E_{z\sim P_z } [f_w (G(z))]+ \lambda_2 E_{\hat{x}\sim P_{\hat{x}}} [(||\nabla_{\hat{x}}   f_w (\hat{x})||_2-1)^2]\\
& 3) \min_{\theta} L_{\text{Gen}} (\theta)=-\min_{\theta} E_{z \sim P_z} [f_w (G(z))].
\end{split}
\label{loss_prop1}
\end{equation}

These losses are trained alternately to optimize different parts of the model. We employ the gradient penalty approach of IWGAN~\cite{gulrajani2017improved} for training the discriminator. In the gradient penalty term, we need to calculate the gradient norm of random samples $\hat{x}\sim P_{\hat{x}}$. According to the proposal in \cite{gulrajani2017improved}, these random samples can be obtained by sampling uniformly along the line connecting pairs of generated and real data samples: 
\begin{equation}
[\hat{x}\sim P_{\hat{x}}] \leftarrow \alpha ~ [x\sim P_{x}] + (1-\alpha)~ [x_{gen}\sim G(z)]   
\end{equation}
The complete training algorithm is described in~\ref{alg:AE-GAN}.

\begin{algorithm}
\caption{TextKD-GAN for text generation.} 
\label{alg:AE-GAN}
\begin{algorithmic}[1]
\Require{The Adam hyperparameters $\alpha$, $\beta_1$, $\beta_2$, the batch size $m$. Initial AE parameters (encoder ($\phi_0$), decoder $\psi_0$), discriminator parameters $w_0$ and initial generator parameters $\theta_0$}

\For {\text{number of training iterations}}
\Statex \hspace{0.5cm} \textbf{AE Training:}
\State {Sample $ \{x^{(i)}\}_{i=1}^m \sim P_x$ and compute code-vectors $c^i = enc_\phi(x^i)$ \\
\hspace{0.5cm} and reconstructed text $\{\tilde{x}^i\}_{i=1}^m$. \\ \hspace{0.5cm} Backpropagate reconstruction loss $L_{AE} (\phi,\psi)$. \\ \hspace{0.5cm}Update with $(\phi,\psi)\gets Adam(L_{AE} (\phi,\psi), \alpha, \beta_1, \beta_2) $}.   
\Statex \hspace{0.5cm}\textbf{Train the discriminator:}
\For {k times}:
\State {Sample $\{x^{(i)}\}_{i=1}^m \sim P_x$ and Sample $\{z^{(i)} \}_{i=1}^m \sim N(0,I)$. \\ \hspace{1cm} Compute generated text $\{x_{gen}^{(i)} \}_{i=1}^m \sim G(z)$ \\ \hspace{1cm} Backpropagate discriminator loss $L_{discriminator}(w)$ . \\ \hspace{1cm} Update with $w\gets Adam(L_{discriminator}(w), \alpha, \beta_1, \beta_2) $}.       
\EndFor $\textbf{end for}$
\Statex	\hspace{0.5cm}\textbf{Train the generator:}

\State {Sample $\{x^{(i)}\}_{i=1}^m \sim P_x$ and Sample $\{z^{(i)} \}_{i=1}^m \sim N(0,I)$. \\ \hspace{0.5cm}Compute generated text $\{x_{gen}^{(i)}\}_{i=1}^m \sim G(z)$ \\ \hspace{0.5cm}Backpropagate generator loss $L_{Gen}(\theta)$. \\ \hspace{0.5cm} Update with $\theta\gets Adam(L_{Gen}(\theta), \alpha, \beta_1, \beta_2) $}.
\EndFor $\textbf{end for}$

\end{algorithmic}
\end{algorithm}

\section{Experiments}
\label{results}
\subsection{Dataset and Experimental Setup}
We carried out our experiments on two different datasets: Google 1 billion benchmark language modeling data\footnote{http://www.statmt.org/lm-benchmark/} and the Stanford Natural Language Inference (SNLI) corpus\footnote{https://nlp.stanford.edu/projects/snli/}. Our text generation is performed at character level with a sentence length of 32. For the Google dataset, we used the first 1 million sentences and extract the most frequent 100 characters to build our vocabulary. For the SNLI dataset, we used the entire preprocessed training data~\footnote{https://github.com/aboev/arae-tf/tree/master/data\_snli}, which contains 714667 sentences in total and the built vocabulary has 86 characters. We train the AE using one layer with 512 LSTM cells~\cite{Hochreiter1997} for both the encoder and the decoder. We train the autoencoder using Adam optimizer with learning rate 0.001, $\beta_1$= 0.9, and $\beta_2$= 0.9. For decoding, the output from the previous time step is used as the input to the next time step.  The hidden code $c$ is also used as an additional input at each time step of decoding. The greedy search approach is applied to get the best output~\cite{kim2017adversarially}. We keep the same CNN-based generator and discriminator with residual blocks as in~\cite{gulrajani2017improved}. The discriminator is trained for 5 times for 1 GAN generator iteration. We train the generator and the discriminator using Adam optimizer with learning rate 0.0001,
$\beta_1$= 0.5, and $\beta_2$= 0.9.


We use the \textit{BLEU-N} score to evaluate our techniques. \textit{BLEU-N} score is calculated according to the following equation~\cite{liu2016not,cer2010best,Papineni2002}: 
\begin{equation}
\textit{BLEU-N} = BP \cdot \text{exp}(\sum_{n=1}^N w_n \text{log} (p_n) )\\
\end{equation}  
where $p_n$ is the probability of $n$-gram and $w_n = \frac{1}{n}$. We calculate BLEU-n scores
for n-grams without a brevity penalty~\cite{Zhu2018}. 
We train all the models for 200000 iterations and the results with the best \textit{BLEU-N} scores in the generated texts are reported. To calculate the \textit{BLEU-N} scores, we generate ten batches of sentences as candidate texts, i.e. 640 sentences (32-character sentences) and use the entire test set as reference texts.


\subsection{Experimental Results}
The results of the experiments are depicted in Table~\ref{tab1} and ~\ref{tab2}. 
As seen in these tables, the proposed TextKD-GAN approach yields significant improvements in terms of \textit{BLEU-2}, \textit{BLEU-3} and \textit{BLEU-4} scores over the IWGAN~\cite{gulrajani2017improved}, and the ARAE~\cite{kim2017adversarially} approaches. Therefore, softened smooth output of the decoder can be more useful to learn better discriminator than the traditional one-hot representation.
Moreover, we can see the lower \textit{BLEU}-scores and less improvement for the Google dataset compared to the SNLI dataset. The reason might be the sentences in the Google dataset are more diverse and complicated. Finally, note that the text-based one-hot discrimination in IWGAN and our proposed method are better than the traditional code-based ARAE technique~\cite{kim2017adversarially}.


\begin{table}[!htb]
  \caption{Results of the \textit{BLEU-N} scores using 1 million sentences from 1 billion Google dataset}
  \label{tab1}
  \centering
  \begin{tabular}{llll}
    \toprule
    Model       &  \textit{BLEU-2} & \textit{BLEU-3} & \textit{BLEU-4}   \\
    \midrule
    IWGAN  & 0.50   & 0.27   & 0.11\\
    ARAE                 &     0.13      &      0.02   &   0.00   \\
    \bottomrule
    TextKD-GAN & \textbf{0.51}  & \textbf{0.29} & \textbf{0.13}\\
    \bottomrule
  \end{tabular}
\end{table}

\begin{table}[!htb]
  \caption{Results of the \textit{BLEU-N} scores using SNLI dataset}
  \label{tab2}
  \centering
  \begin{tabular}{llll}
    \toprule
    Model        & \textit{BLEU-2} & \textit{BLEU-3} & \textit{BLEU-4}   \\
    \midrule
    IWGAN  & 0.57   & 0.44   & 0.30 \\
    ARAE                 &     0.37      &    0.27     &  0.17    \\
    \bottomrule
    TextKD-GAN  & \textbf{0.62}  & \textbf{0.50} & \textbf{0.38} \\
    \bottomrule
  \end{tabular}
\end{table}

Some examples of generated text from the SNLI experiment are listed in Table~\ref{tab3}. As seen, the generated text by the proposed TextKD-GAN approach is more meaningful and contains more correct words compared to that of IWGAN~\cite{gulrajani2017improved}. 
\begin{table}[!htb]
  \caption{Example generated sentences with model trained using SNLI dataset}
  \label{tab3}
  \centering
  \begin{tabular}{|l|l|}
    \hline
           \hspace{1.75cm}IWGAN       &       \hspace{1.75cm}{TextKD-GAN}   \\ 
    \hline
     The people are laying in angold & Two people are standing on the s\\ 
A man is walting on the beach & A woman is standing on a bench .\\
A man is looking af tre walk aud & People have a ride with the comp\\
A man standing on the beach  & A woman is sleeping at the brick\\
The man is standing is standing & Four people eating food .\\
A man is looking af tre walk aud & The dog is in the main near the \\
The man is in a party . & A black man is going to down the\\
Two members are walking in a hal & These people are looking at the \\
A boy is playing sitting . & the people are running at some l
 \\
\hline
  \end{tabular}
\end{table}
\begin{figure*}[!htb]
\centering
\begin{subfigure}[t]{0.45\textwidth}
\centering
\includegraphics[scale=0.35]{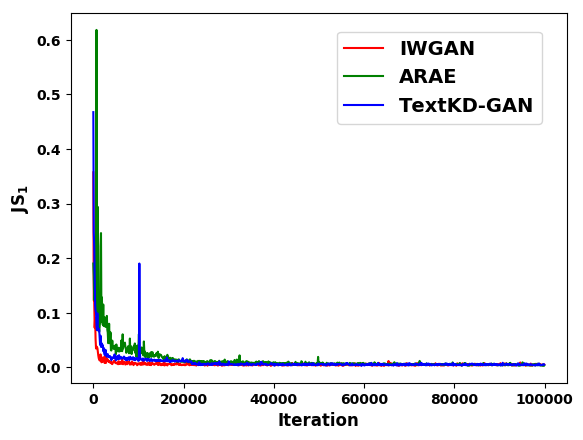}
\caption{}
\end{subfigure}
\begin{subfigure}[t]{0.45\textwidth}
\centering
\includegraphics[scale=0.35]{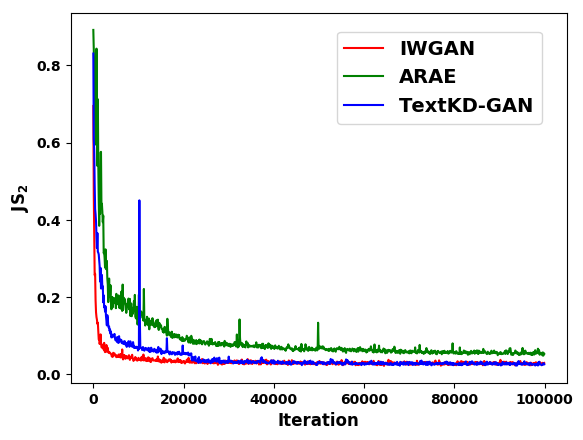}
\caption{}
\end{subfigure}
\begin{subfigure}[t]{0.45\textwidth}
\centering
\includegraphics[scale=0.35]{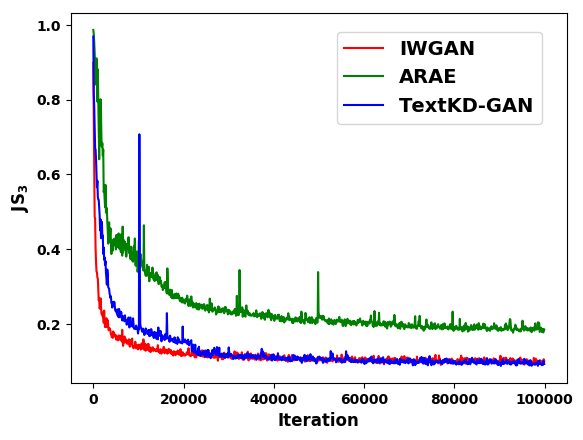}
\caption{}
\end{subfigure}
\begin{subfigure}[t]{0.45\textwidth}
\centering
\includegraphics[scale=0.35]{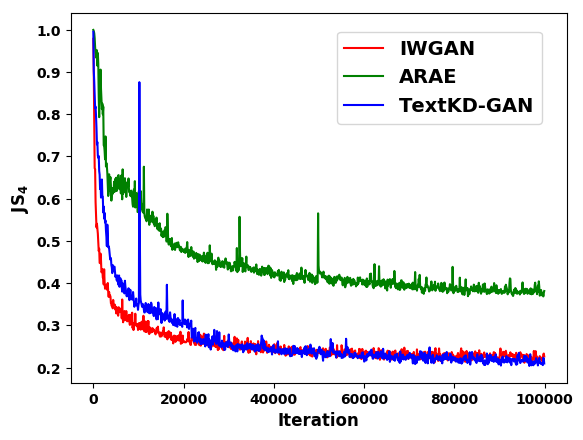}
\caption{}
\end{subfigure}
\caption{Jensen-Shannon distance (JSD) between the generated and training sentences $n$-grams derived from SNLI experiments. a) $js_1$, b) $js_2$, c) $js_3$, and d) $js_4$ represent the JSD for 1, 2, 3, and 4-grams respectively}
\label{js}
\end{figure*}

We also provide the training curves of Jensen-Shannon distances (JSD)  between the $n$-grams of the generated sentences and that of the training (real) ones in Figure~\ref{js}. The distances are derived from SNLI experiments and calculated as in~\cite{gulrajani2017improved}. That is by calculating the log-probabilities of the $n$-grams of the generated and the real sentences. As depicted in the figure, the TextKD-GAN approach further minimizes the JSD compared to the literature methods~\cite{gulrajani2017improved,kim2017adversarially}. In conclusion, our approach learns a more powerful discriminator,  which in turn generates the data distribution close to the real data distribution.

\subsection{Discussion}
The results of our experiment shows the superiority of our TextKD-GAN method over other conventional GAN-based techniques. We compared our technique with those GAN-based generators which does not need pre-training. This explains why we have not included the RL-based techniques in the results. We showed the power of the continuous smooth representations over the well-known tricks to work around the discontinuity of text for GANs. Using AEs in TextKD-GAN adds another important dimension to our technique which is the latent space, which can be modeled and exploited as a separate signal for discriminating the generated text from the real data. It is worth mentioning that our observations during the experiments show training text-based generators is much easier than training the code-based techniques such as ARAE. Moreover, we observed that the gradient penalty term plays a significant part in terms of reducing the mode-collapse from the generated text of GAN. Furthermore, in this work, we focused on character-based techniques; however, TextKD-GAN is applicable to the word-based settings as well. 
Bear in mind that pure GAN-based text generation techniques are still in a newborn stage and they are not very powerful in terms of learning semantics of complex datasets and large sentences. This might be because of lack of capacity of capturing the long-term information using CNN networks. To address this problem, RL can be employed to empower these pure GAN-based techniques such as TextKD-GAN as a next step .   

\section{Conclusion and Future Work}
\label{conclusion}
In this work, we introduced TextKD-GAN as a new solution using knowledge distillation for the main bottleneck of using GAN for generating text, which is the discontinuity of text. Our solution is based on an AE (Teacher) to derive a continuous smooth representation of the real text.  This smooth representation is distilled to the GAN discriminator instead of the conventional one-hot representation. We demonstrated the rationale behind this approach, which is to make the discrimination task of the discriminator between the real and generated texts more difficult and consequently providing a richer signal to the generator. At the time of training, the TextKD-GAN generator (Student) would try to learn the manifold of the smooth representation, which can later on be mapped to the real data distribution by applying the argmax operator. We evaluated TextKD-GAN over two benchmark datasets using the \textit{BLEU-N} scores, JSD measures, and quality of the output generated text. The results showed that the proposed TextKD-GAN approach outperforms the traditional GAN-based text generation methods which does not need pre-training such as IWGAN and ARAE. Finally, We summarize our plan for future work in the following: 
\begin{enumerate}

\item We evaluated TextKD-GAN in a character-based level. However, the performance of our approach in word-based level needs to be investigated.

\item Current TextKD-GAN is implemented with a CNN-based generator. We might be able to improve TextKD-GAN by using RNN-based generators. 
   
\item TextKD-GAN is a core technique for text generation and similar to other pure GAN-based techniques, it is not very powerful in generating long sentences. RL can be used as a tool to accommodate this weakness.    
\end{enumerate}

\bibliographystyle{splncs04}
\bibliography{mybibliography}




\end{document}